\documentclass[sigconf]{acmart}

\usepackage[noend]{algorithmic}
\usepackage{algorithm}
\usepackage{multirow}
\usepackage{flushend}
\usepackage{amsmath,amssymb}
\usepackage{mathtools}
\usepackage[acronym]{glossaries}
\usepackage{units}
\usepackage{booktabs}
\usepackage[inline]{enumitem} 
\usepackage{subcaption}
\usepackage{bm} 
\usepackage{amsmath,amssymb}
\usepackage{dirtytalk}
\usepackage{tikz}
\usepackage{hyperref}
\usetikzlibrary{bayesnet}
\usetikzlibrary{positioning}

\newcommand{\bL}{\boldsymbol{L}}

\DeclareMathOperator*{\E}{\mathbb{E}}

\newcommand{\bepsilon}{\boldsymbol{\epsilon}}

\newcommand{\bmu}{\boldsymbol{\mu}}
\newcommand{\bSigma}{\boldsymbol{\Sigma}}

\newcommand{\bPsi}{\boldsymbol{\Psi}}

\newcommand{\bomega}{\boldsymbol{\omega}}

\newcommand{\bI}{\boldsymbol{I}}
\newcommand{\bzero}{\boldsymbol{0}}
\newcommand{\brho}{\boldsymbol{\rho}}

\newcommand{\bv}{\boldsymbol{v}}

\newcommand{\bbeta}{\boldsymbol{\beta}}
\newcommand{\bkappa}{\boldsymbol{\kappa}}
\newcommand{\bzeta}{\boldsymbol{\zeta}}
\newcommand{\bR}{\boldsymbol{R}}
\newcommand{\bsigma}{\boldsymbol{\sigma}}

\copyrightyear{2020} 
\acmYear{2020} 
\setcopyright{acmcopyright}
\acmConference[KDD '20] {26th ACM SIGKDD Conference on Knowledge Discovery and Data Mining}{August 23--27, 2020}{Virtual Event, USA}
\acmBooktitle{26th ACM SIGKDD Conference on Knowledge Discovery and Data Mining (KDD '20), August 23--27, 2020, Virtual Event, USA}
\acmPrice{15.00}
\acmDOI{10.1145/3394486.3403121}
\acmISBN{978-1-4503-7998-4/20/08} 

\begin{document}
\fancyhead{}

\title{BLOB : A Probabilistic Model for Recommendation that Combines Organic and Bandit Signals}

\author{Otmane Sakhi}
\affiliation{
\institution{Criteo AI Lab}
\city{Paris}
\country{France}
}
\email{o.sakhi@criteo.com}

\author{Stephen Bonner}
\affiliation{
  \institution{Department of Computer Science, Durham University}
  \city{Durham}
  \country{UK}
}
\email{s.a.r.bonner@durham.ac.uk}

\author{David Rohde}
\affiliation{
  \institution{Criteo AI Lab}
  \city{Paris}
  \country{France}
}
\email{d.rohde@criteo.com}

\author{Flavian Vasile}
\affiliation{
  \institution{Criteo AI Lab}
  \city{Paris}
  \country{France}
}
\email{f.vasile@criteo.com}

\renewcommand{\shortauthors}{O. Sakhi et al.}

\begin{abstract}
  A common task for recommender systems is to build a profile of the interests of a user from items in their browsing history and later to recommend items to the user from the same catalog. The users' behavior consists of two parts: the sequence of items that they viewed without intervention (the organic part) and the sequences of items recommended to them and their outcome (the bandit part).  
  In this paper, we propose \emph{Bayesian Latent Organic Bandit model (BLOB)},  a probabilistic approach to combine the `organic' and `bandit' signals in order to improve the estimation of recommendation quality.  The bandit signal is valuable as it gives direct feedback of recommendation performance, but the signal quality is very uneven, as it is highly concentrated on the recommendations deemed optimal by the past version of the recommender system.  In contrast, the organic signal is typically strong and covers most items, but is not always relevant to the recommendation task. In order to leverage the organic signal to efficiently learn the bandit signal in a Bayesian model we identify three fundamental types of distances, namely action-history, action-action and history-history distances.  We implement a scalable  approximation of the full model using variational auto-encoders and the local re-paramerization trick. We show using extensive simulation studies that our method out-performs or matches the value of both state-of-the-art organic-based recommendation algorithms, and of bandit-based methods (both value and policy-based) both in organic and bandit-rich environments. 
\end{abstract}

  
  
\maketitle

\section{Introduction}
The recommender systems literature is somewhat bifurcated into two distinct branches.  One branch concerns analysing logs of organic user sessions where similar items co-occur \cite{adomavicius2005toward,koren2015advances,hidasi2018recurrent,liang2018variational}.  A distinguishing feature of this research is that it focuses on logs of organic user sessions where users view variable numbers of (usually) related items in a shopping session.

A second branch of research explicitly (and entirely) focuses on the logs of the recommender system using the history of successful and unsuccessful recommendations in order to discover a good recommender system policy.  This branch uses off policy learning in order to discover new policies with good actions \cite{Bottou2013,beygelzimer2009offset,swaminathan2015batch}.  This work is distinguished by its use of recommender system logs for training and its  anonymous feature vector (usually called the context).

The purpose of this paper is twofold.  Firstly, we pose a simple yet powerful model that combines these two distinct data sources in order to efficiently learn good recommendation policies.  Secondly, we develop a fully probabilistic approach to recommendation and outline its benefits and consequences.  The probabilistic formulation gives insights into user embedding creation and the alternative frameworks of value and policy learning.  

The remainder of the paper is structured as follows: In Section 2 we introduce our probabilistic model of organic and bandit behaviour and discuss its properties.  In Section 3 we describe the training of the model.  In section 4 we apply our model to the RecoGym simulator \cite{rohde2018recogym,Jeunen2020} and present results.  Concluding remarks are made in Section 5.
\section{Probabilistic model of organic and bandit sessions}
We develop a simple probabilistic model that allows us to build a representation of a user from a variable length organic sequence of items and then predict accurately how probable the user is to respond positively to each recommendation in the catalog.

Throughout this paper, we will make use of the notation introduced in Table \ref{tab:notation}. We use $u$ to denote a user or a session, we use $t$ time to denote sequential time and $v$ to denote which product they viewed from $1$ to $P$ where $P$ is the number of products.  User $u$ will also be given some recommendations (or actions) $a_{u,1},...,a_{u,n}$ again which can take values from $1$ to $P$ and we will observe a reward (or a click) for each of these recommendations $c_{u,1},...,c_{u,n}$.  The organic part of the session are the items the user views without any encouragement from the recommender system i.e. $v_{u,1},...v_{u,T_u}$, the bandit part of the session refers to the recommender system log: $a_{u,1},...,a_{u,n_u}; c_{u,1},...,c_{u,n_u}$. Thus, the size of the organic dataset is U, the number of users, and the bandit dataset size is $\sum_u n_u = N$. We drop the $u$ subscript and treat the bandit dataset as records with $n \in [1,...,N]$.

In our model, the user's interest is described by a $K$ dimensional variable $\bomega_{u}$ which can be interpreted as the user's interest in $K$ topics. We then assume the following generative process for the organic views in each session:
\[
\bomega_u \sim \mathcal{N}(\bzero_K, \bI_K), \hspace{2mm} v_{u,1}, .., v_{u,T_u} \sim {\rm categorical}({\rm softmax} (\bPsi \bomega_u + \brho) ) 
\]

\noindent
The organic embedding matrix $\bPsi$ is $P \times K$ and represents information about how items correlate in a users session organically (i.e. without any intervention from the recommender system).  The $P$ dimensional vector $\brho$ is related to the items organic popularity.

Once this session is generated a recommendation or actions is made to user $u$ denoted $a_u$ and a reward or click will be observed $c_u$.
\[
c_u | a_u ,\bbeta,\bomega,\bkappa \sim {\rm Bernoulli}\{ {\rm sigmoid}(\bbeta_{a_u} \bomega_u + \bkappa_{a_u}) \}
\]

\noindent
The bandit embedding matrix $\bbeta$ is $P \times K$ and represents information about how to personalise recommendations to a user $u$ with a latent user representation $\bomega_u$.

\noindent
The organic behavior is parameterized by $\bPsi,\brho$ and the bandit behavior is parameterized $\bbeta,\bkappa$ in order to relate the two we use the following matrix variate prior distribution of $\bbeta$:
\[
\bbeta|\bPsi \sim \mathcal{MN}(s^+(w_a) \bPsi, s^+(w_b) \bPsi \bPsi^T, s^+(w_b) \frac{1}{P} \bPsi^T \bPsi).
\]

\noindent
Where $\mathcal{MN}(\cdot)$ is the matrix variate normal distribution\footnote{The matrix normal distribution can be defined by its connection to the multivariate normal.  If $\bbeta \sim \mathcal{MN}(\boldsymbol{M},\boldsymbol{R},\boldsymbol{S})$, where mean matrix $\boldsymbol{M}$ is $M \times N$, and $\boldsymbol{R}$ is $M \times M$ and $\boldsymbol{S}$ is $N \times N$ - then:
 ${\rm vec}(\bbeta) \sim \mathcal{N}({\rm vec}(\boldsymbol{M}),\boldsymbol{R} \otimes \boldsymbol{S})$.  In this way the matrix variate normal has a more compact and restricted representation of the co-variance than the matrix variate normal.  Here $\otimes$ denotes the Kronecker product.}  We will show how each of the three terms in the matrix variate normal allow us to include in our model one of the three fundamental differences of recommendation.  The softplus function is defined:
\[
s^+(w) = \log \{1 + \exp(w)\}.
\]

\noindent
We also put a prior on  $\bkappa$ which is $P \times 1$:
\[
\bkappa \sim \mathcal{N}(w_c, \bI_P \sigma^2_{\kappa}).
\]

\noindent
The hyper-parameters $w_a,w_b,w_c$ are also given normal priors:
\[
w_a \sim \mathcal{N}(\mu_{w_{a_0} },\sigma^2_{w_{a_0}}), \hspace{0.15cm} w_b \sim \mathcal{N}(\mu_{w_{b_0} },\sigma^2_{w_{b_0} }), \hspace{0.15cm} w_c \sim \mathcal{N}(\mu_{w_{c_0} },\sigma^2_{w_{c_0}} ).
\]

\begin{figure}
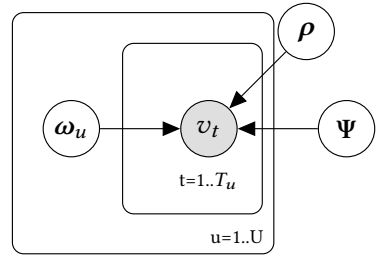

  \centering
  \resizebox{5cm}{!}{
    \tikz{ %
      \node[obs] (V_t) {$v_t$} ; %
      \node[latent, right=of V_t] (psi) {$\bPsi$} ; %
      \node[latent, above right=of V_t] (rho) {$\brho$} ; %
      \node[latent, left=of V_t] (omega) {$\bomega_u$} ; %
      \edge {psi} {V_t} ; %
      \edge {rho} {V_t} ; %
      \edge {omega} {V_t} ; %
      \plate[inner sep=0.5cm, xshift=-0.2cm, yshift=0.2cm] {plate1} {(V_t)} {t=1..$T_u$};
      \plate[inner sep=0.25cm, xshift=-0.12cm, yshift=0.12cm] {plate2} {(omega)  (plate1)} {u=1..U};
    
    }
  }
  \caption{A graphical model of the organic behavior.}
  \label{fig:pgm1}  
\end{figure}

\begin{figure}
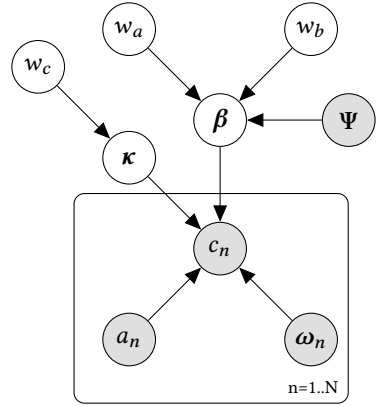

  \centering
  \resizebox{5cm}{!}{
    \tikz{ %
      \node[obs] (C_n) {$c_n$} ; %
      \node[obs, below left=of C_n] (a_n) {$a_n$} ; %
      \node[obs, below right=of C_n] (omega_n) {$\bomega_n$} ;
      \edge {a_n} {C_n} ; %
      \edge {omega_n} {C_n} ; %
      \plate[inner sep=0.25cm, xshift=-0.12cm, yshift=0.12cm] {plate1} {(C_n) (a_n) (omega_n)} {n=1..N};
      \node[latent, above=of C_n] (beta) {$\bbeta$} ; %
      \node[obs, right=of beta] (psi) {$\bPsi$} ;
      \node[latent, above left=of C_n] (kappa) {$\bkappa$} ; %
      \node[latent, above left=of kappa] (w_c) {$w_c$} ; %
      \node[latent, above left=of beta] (w_a) {$w_a$} ; %
      \node[latent, above right=of beta] (w_b) {$w_b$} ; %
      \edge {psi}{beta};%
      \edge {beta} {C_n} ; %
      \edge {kappa}{C_n};
      \edge {w_a}{beta};
      \edge {w_b}{beta};
      \edge {w_c}{kappa};
      
    }
  }
  \caption{A graphical model of the bandit behavior.}
  \label{fig:pgm2}  
\end{figure}

In this paper we will mostly consider the organic and bandit behavior as separate but related processes.  A graphical model defining the organic portion of the model is given in Figure \ref{fig:pgm1}.  This  graphical model has a similar structure to the latent Dirichlet Allocation model (LDA) \cite{blei2003latent}, the difference being that where we model $v \sim {\rm categorical}\{ \rm softmax(\bPsi \bomega + \brho)\}$, LDA uses  $v \sim {\rm categorical}(\bPsi \bomega)$ putting simplex constraints on $\bPsi$ and $\bomega$, similarly correlated topic models \cite{lafferty2006correlated} use $v \sim {\rm categorical} \{\bPsi {\rm softmax}(\bomega) \}$ where the simplex constraint is only on $\bPsi$.  This model can also be viewed as a linear version of the Multi-VAE \cite{liang2018variational}.

We will show that using variational autoencoders with the re-parameterization trick is an effective way to train  the organic model.

The approach developed in this paper takes the organic model and estimates $\bPsi$ by maximum likelihood and $\bomega$ by posteior mean (denoted $\hat{\bomega}$) and then treats $\bPsi$ and $\hat{\omega}$ as observed in the bandit model.  The graphical model is shown in Figure \ref{fig:pgm2}.  In this probabilistic model we will develop full Bayesian inference of the $\bbeta$, $\bkappa$, $w_a$, $w_b$ and $w_c$. This is important because the bandit signal is very uneven.  Lots of information is available on past actions that the previous recommender system favoured and little information or no information is available on many other actions, meaning that the posterior is tight in some regions but broad and highly influenced by the prior in others.  We use variational approximations and the local re-parameterization trick in order to capture this complex structure.

We refer to the organic only component of the model as \textbf{BLO} (Bayesian Latent Organic) model (we apply maximum likelihood to $\bPsi,\brho$ and integrate $\bomega$).  The full model is referred to as \textbf{BLOB} (Bayesian Latent Organic Bandit Model).

\subsection{Intuition for the model}

The model presented embodies a fundamental implicit assumption in the traditional recommendation system, the assumption that auto-completion of a session results in good recommendations being made.  This is one of the three fundamental distances of recommendation, the action-history distance.

\subsubsection{The implicit assumption in traditional recommendation: good recommendations are (usually) similar to the items in the user's history}

Algorithms in the recommendation literature look at items in a user's history and attempt to predict the final element in this session.  The fraction of times that the predicted item is within the top K items in a held out data set is a key metric called precision@K that measures a models ability to ``auto-complete'' a users behavior.  The organic performance is therefore computed:
\[
P(v_{u,T_u}|v_{u,1},..,v_{u,T_u-1}).
\]

\noindent
Metrics such as NDCG, recall@K or log likelihood are computed on this auto-completion task.

However auto-completion is not the same as recommendation.  In fact to reduce recommendation to auto-completion removes the opportunity for a recommender system to help a user discover new things which arguably is the primary objective of recommendation.  That said, organic data is usually plentiful and this implicit assumption that recommendation as auto-completion certainly has some merit.  We can state this assumption as, if:
\begin{align*}
P(V_{u,T_u}=v_a&|v_{u,1},..,v_{u,T_u-1}) 
> P(V_{u,T_u}=v_b|v_{u,1},..,v_{u,T_u-1})
\end{align*}

\begin{table}
\begin{tabular}{c l r}
  \toprule
  \textbf{Symbol} & \textbf{Dimension} & \textbf{Description} \\ 
 \midrule
 \midrule
 $u$ & Scalar & A given user's id.  \\
 $t$ & Scalar & sequential time. \\
  $P$ & Scalar & Total number of products. \\
  $K$ & Scalar & The size of the embedding. \\
  $v_{u,t}$ & Scalar & Product id for user $u$ at time $t$.\\
  $\bomega_u$ & $K \times 1$& A given user's state.\\
  $\bPsi$ & $P \times K$ & Organic embedding matrix.\\
  $\bPsi_v$ & $1 \times K$ & Organic embedding for $v$.\\
  $\bbeta$ & $P \times K$ & Bandit embedding matrix.\\
  $\bbeta_v$ & $1 \times K$ & Bandit embedding for $v$.\\
  $\brho$ & $P \times 1$ & Item popularity intercept.\\
  $\bkappa$ & $P \times 1$ & Item recommendability intercept.\\
  $T_u$ & Scalar & Session length for $u$.\\
  $N$ & Scalar & The size of the Bandit dataset.\\
  $U$ & Scalar & The number of user sessions.
     \\
  \bottomrule
  \end{tabular}
  \caption{Notations and Definitions} 
  \label{tab:notation}
\end{table}

\noindent
Then item $v_a$ is probably better than item $v_b$ as a recommendation i.e the following holds with high probability:
\begin{align*}
P(c=1|&A=v_a,v_{u,1},..,v_{u,T_u-1}) > P(c=1|A=v_b,v_{u,1},..,v_{u,T_u-1})
\end{align*}

\noindent
Although this relationship often holds, it need not hold in every single instance.  Maybe the user already knows about item $v_a$, maybe the recommendation for $v_a$ is unattractive or maybe the reason the user never visited item $v_b$ is lack of knowledge and it is actually a very valuable recommendation.  We want our recommender system to make use of the organic relationship, but we also want to learn from the logs of the recommender system itself which records if the recommendations that we chose to deliver were successful or not.  This ``bandit feedback'' is in some sense the true arbiter of if a recommendation is good or not, but the bandit signal is usually highly concentrated around what the previous version of the recommendation system judged to be a good recommendation, so it cannot reliably be used over the entire recommendation space.  For example the organic session might contain information that two products (say) rice and a phone are rarely viewed together in the same organic session.  However it probably will not contain many events where a phone is recommended to a user with rice in their history.  If the recommender system is to infer that this is likely a poor recommendation, it must do so through a prior linking the bandit behavior to the organic behavior.

When deployed in a production recommender system the model operates in the following way.  First a posterior over a user embedding is approximately calculated:

\[
P(\bomega_u|v_{u,1}, ..., v_{u,T_u}, \bPsi, \brho) 
\]

\sloppy

A fast variational approximation can be made of $\bomega_u \sim \mathcal{N}(\bmu_{\omega_q}, \bSigma_{\omega_q})$ which gives both a mean and a variance (this can be done using either a variational EM algorithm or a variational autoencoder).  

For our purposes we make the pragmatic compromise that we can summarise the user history with a posterior mean point estimate $\hat{\bomega} = \bmu_{\omega_q}$, this prevents numerical integration of $\bomega_u$ at recommendation time.  Once this compromise is made it also makes sense to train the organic and bandit components separately.  The probability of a click is given by:

\[
P(c|\hat{\bomega},\bbeta,\bkappa,a) = {\rm sigmoid}(\bbeta_a \hat{\bomega} + \bkappa_a)
\]

The recommender system will then choose a recommendation that will optimise this reward (or a combination of reward and exploration - but the explore-exploit dilemma \cite{lattimore2018bandit} is beyond the scope of this paper.  

The organic parameters $\bPsi$ and $\brho$ are not required in order to deliver a recommendation.  They are used only to put a prior on the bandit embeddings.

We note parenthetically that due to the fact that once the user embedding $\hat{\bomega}$ is created the model is linear and we can exploit fast algorithms to quickly find the optimal recommendation over large catalogues \cite{gionis1999similarity,malkov2018efficient}.


\subsubsection{The organic user session}

The organic user session model we propose can be understood in a number of ways.  It can be viewed as a user item matrix factorization where the user has a latent interest in $K$ topics - a discussion of this interpretation is given in the supplementary material.

It can also be viewed as an i.i.d. categorical process with a (usually) low-rank multivariate normal prior.  The prior causes similar items to co-occur in a session with high probability.  Because of this assumptions seeing an item will always make it more likely to be viewed again.  If we had a full rank model the user session would imply the law of large numbers where the next item prediction will converge to the empirical frequency.  In practice the session history is short and the embedding size is much lower than the number of products, but the assumption remains that viewing an item makes  the conditional probability for that same item increase (also and importantly the conditional probability that similar items will be viewed also increases).

This is a relatively strong assumption compared to powerful sequential models such as recurrent neural networks \cite{hidasi2018recurrent} which can model complex sequences. 
The simpler and stronger assumption made by BLO is reasonable in many settings and greatly simplifies learning.

\subsubsection{The bandit session and the three distances in recommendation}

The auto-complete assumptions as embodied in the recommendation research measures the similarity between the recommendation and the items in history.  This is the first similarity or distance, the distance between the history and the action.  The mean of the matrix normal $\bPsi$ embodies this assumption.

The second similarity in recommendation is the similarity between actions.  That is if action $a_1$ and $a_2$ are similar then we expect that the responses to these actions to the same (or similar) users be correlated.  This distance is encoded with the first (low rank) co-variance $\bPsi \bPsi^T$ in the matrix normal prior on $\bbeta$.

The third similarity in recommendation is the similarity between users.  If user $u_1$ and $u_2$ are similar then we expect the response to the same (or similar) action on these users to be correlated.  This distance is encoded with the second co-variance $\bPsi^T \bPsi$ in the matrix normal prior on $\bbeta$.

The effect of the first distance is to seed the recommendation using the organic similarities, the effect of the second and third is to borrow strength allowing the bandit signal to be used more effectively.  Finally the parameters $w_a$ and $w_b$ control the strength of the influence of the first and second distance.  The relative strength of the first distance and the second is an extremely important hyper-parameter.

\subsection{Value vs policy learning}

The method proposed here is a value based method as it learns the value for every action and then can determine a decision rule using unconstrained optimisation.  In this way it differs from alternative methods for learning from bandit feedback that have been recently proposed \cite{Bottou2013,beygelzimer2009offset,swaminathan2015batch} which use policy learning.

Bayesian methods are inherently value based and bring the benefit of being able to synthesis data sources such as organic and bandit, they also produce uncertainty that is useful for explore-exploit strategies such as upper confidence bound and Thompson sampling \cite{lattimore2018bandit}.  From a purely statistical point of view principles such as the conditionality and the likelihood principle actually forbid the use of the propensity score \cite{berger1988likelihood,hernan2010causal}.  Given that training on bandit feedback is sometimes considered to be synonymous with using the inverse propensity score (IPS) it is worth reviewing some advantages of Bayesian value based methods.

It has been shown in \cite{ritov2014bayesian}, that under regularity conditions that apply in the recommendation case, the Bernstein-von Mises theorem applies, and that the Bayesian estimator is efficient $\sqrt{n}$ consistent and necessarily better than the IPS (or Horvitz-Thompson) estimator\footnote{They additionally show that IPS based methods can have better frequentist properties than Bayesian estimators when these regulatory conditions break down.}.  However note that a real recommender system log will be of sufficient dimensionality that even with terabytes of logs asymptotic theory is usually not relevant (i.e. priors will have real impacts).

It is also sometimes argued that the IPS score is necessary to apply in counterfactual settings due to the domain shift which occurs in causal settings \cite{johansson2016learning}.  However this argument does not apply when the model has enough capacity to accurately predict the value everywhere \cite{Storkey2009} and there is no need to constrain capacity to reduce estimator variance when applying Bayesian methods \cite{neal2012bayesian}.  It seems that some of the positive aspects of value based methods have been overlooked due to criticisms that apply only in the non-Bayesian case.

Policy learning also suffers from some draw backs.  Policy learning extends the principle of Statistical Learning Theory (SLT) to the counterfactual setting.  The idea of SLT is that a decision rule is fit to the historical data from a constrained set. If a decision rule from a restricted set has good performance (low risk) then it is likely to also have low risk on out of sample data \cite{vapnik2013nature}.  These analyses are based upon treating empirical risk or counterfactual risk as a statistic, but these are highly non-sufficient statistics and there is no ability to order decision rules that have the same empirical risk even when away from the data they are very different.  The theory is heavily based on having a restricted set of decision rules, but restricting the set might exclude good decisions.  Value based methods make no such restriction.

Extending SLT to the counterfactual setting requires some additional ideas because the consequences of decisions the new policy will make are not available.  IPS based methods have been a recent research focus that extend the empirical risk minimisation to the counterfactual setting.  Technical challenges are being addressed such as the fact that the variance of the decision rule can vary depending on how much it differs from the historical logging policy \cite{Swaminathan2015JMLR}.  As well as the problem of propensity overfitting i.e. decision rules can achieve an estimated reward of 0 by avoiding past decisions (0 might be good or bad depending on how the reward is defined) causing decision rules either to cling to the old policy or to be driven away from it\footnote{The self normalized importance sampling variant of IPS is one proposal to remove this sensitivity to the definition of the reward\cite{schnabel2016recommendations}}.  It is usually considered a better heuristic for the new policy to cling to the old one.



One simple method to control variance is to cap large weights \cite{Bottou2013} (necessarily associated with actions that are different to the logging policy). This method controls the bias-variance trade-off.  Another method that more explicitly discourages deviation from the logging policy is to apply variance penalization \cite{Swaminathan2015JMLR} here rather than optimizing the conterfactual risk directly a penalized term is instead optimized, this penalization naturally goes up if the recommendations are rare under the logging policy (and hence have a high IPS weight).  


Many of the standard policy learning settings \footnote{This includes having reward positive and no-reward zero, capping and variance penalization}  have the property that the learnt policy will only deviate from the preferred decision of the logging policy in the face of considerable evidence.  This is a good heuristic in cases where the logging policy is good, but can be a problem in other situations.  

The potential strength of policy based approaches is due to the fact they do not use a model and they focus directly on the decision rule focusing optimisation and capacity on the parts of the problem that matters most.  Bayesian value based methods cannot do this because the modelling step is made before and separately to the decision making step.

\section{Model Training}
\subsection{Organic session training: learning the organic embeddings}

The log likelihood of the organic model has the form:
\begin{align*}
  & \log  p(v_1,..,v_T,\bomega_u|\bPsi) = \left( \sum_t^T \bPsi_{v_t}
            \bomega_u + \brho_{v_t}\right)  \\
&            -T \log\{ \sum_p^P \exp( \bPsi_p \bomega_u  +\brho_p)\} + \log p(\bomega_u) 
\end{align*}  

\noindent
As the posterior on $\omega$ is intractable, we use a normal distribution $\bomega_u \sim \mathcal{N}(\bmu_{q_\omega},\bSigma_{q_\omega})$ to approximate it, we get a variational lower bound of the form:
\begin{align*}
& \mathcal{L} = \E_{q(\bomega_u)}[\log ~   p(v_1,..,v_T,\bomega_u|\bPsi) -\log q(\bomega_u)] = \\
&\left( \sum_t^T \bPsi_{v_t}
            \bmu_{q_\omega} + \brho_{v_t}\right)  -T \E_{q(\bomega_u)} [\log \{\sum_p^P \exp( \bPsi_p \bomega_u  +\brho_p)\}]\\
&  - \rm KL (q(\bomega_u)|p(\bomega_u)).
\end{align*}  

\noindent
Where KL is a closed form KL divergence between the variational posterior and the prior (a multivariate standard normal distribution). We see that there is a problematic term associated with the denominator of the softmax.  We use the re-parameterization trick \cite{kingma2013auto} to overcome this term.  It is also possible to use the Bouchard bound (which also enables an EM algorithm) and the log concave bound, both bounds can alleviate computational issues associated with the softmax sum \cite{bouchard2007efficient}, details of these lower bounds and the EM and simulated EM algorithm are given in the supplementary material.

\subsubsection{Re-parameterization Trick}

An effective approach to computing expectations with respect to the denominator of the softmax is to use the re-parameterization trick \cite{kingma2013auto}, which allows us to take a  sample of $\bomega$ from the variational distribution and compute a noisy derivative of the lower bound.  Within each iteration we proceed by simulating: $\bepsilon^{(s)} \sim \mathcal{N}(\boldsymbol{0}_K, \bI_K)$,
and then computing: $\bomega^{(s)} = L_{\bSigma_{q_\omega}} \bepsilon^{(s)} + \bmu_{q_\omega}$.
Where $\bL_{\bSigma_{q_\omega}} \bL_{\bSigma_{q_\omega}}^T = \bSigma_{q_\omega}$, we can then optimize the noisy lower bound:
\begin{align*}
   \mathcal{L}_{MC} = &
  \left( \sum_t^T \bPsi_{v_t}
              \bmu_{q_\omega} + \brho_{v_t}\right) - \rm KL(q(\bomega_u)|p(\bomega_u)) \\
              & -T \log[ \sum_p^P \exp\{ \bPsi_p (L_{\bSigma_{q_\omega}} \bepsilon^{(s)} + \bmu_{q_\omega}  )  +\brho_p\}]
\end{align*}  

\noindent
Often $\bSigma_{q_\omega}$ is taken to be diagonal which makes computing $\bL_{\bSigma_{q_\omega}}$ simply an element-wise square root.

A naive application of the algorithm discussed so far would have the number of variational parameters $\bmu_{q_\omega},\bSigma_{q_\omega}$ growing with the number of user sessions. We propose instead to limit the number of parameters by the use of a variational auto-encoder \cite{kingma2013auto}.  This involves using a flexible function and optimizing it to do the job of the EM algorithm i.e.

\[
\bmu_{q_\omega}, ~ \bSigma_{q_\omega} = f_\Xi(v_1,...v_{T}),
\]

\noindent
Where any function e.g. a deep net can be used for $f_\Xi(\cdot)$ such as a deep or shallow neural network.

\subsection{Bandit session training: learning the bandit embeddings}

For every user we compute: $\hat{\bomega}_u = f(\bv_u)$ (uncertainty over $\bomega_u$ is ignored and a point estimate taken).  The hierarchical model has the form:
\[
w_a \sim \mathcal{N}(\mu_{0_{w_a}},\sigma^2_{0_{w_a}}), ~ ~ w_b \sim \mathcal{N}(\mu_{0_{w_b}},\sigma^2_{0_{w_b}}), ~~ w_c \sim \mathcal{N}(\mu_{0_{w_c}},\sigma^2_{0_{w_c}})
\]
\vspace{-2mm}
\[
\bkappa' \sim \mathcal{N}(\bzero_P, \sigma_{\kappa_0}^2 \bI_P), \hspace{1cm} \bkappa = \bkappa' + w_c
\]
\vspace{-2mm}
\[
\bbeta|\bPsi, w_a, w_b \sim \mathcal{MN}(s^+(w_a) \bPsi, s^+(w_b) \bPsi \bPsi^T, s^+(w_b) \frac{1}{P} \bPsi^T \bPsi  )
\]
\vspace{-2mm}
\[
c_n | a_n ,\bbeta,\bomega,\bkappa \sim {\rm Bernoulli}\{ {\rm sigmoid}(\bbeta_{a_n} \bomega_n + \bkappa_{a_n}) \}.
\]

\noindent
While $\beta$ is a [P x K] random variable, we can leverage its low rank covariance matrix to transform the problem to infering a posterior on a [K x K] random variable. This reduces dramatically the training time as P, the size of the catalog items is usually very large compared with K. The low rank alternative parameterization of this distribution can be set as follows. Let:
\[
\bzeta \sim \mathcal{MN}(\bzero_{K,K}, \bI_K, \bI_K).
\]

\noindent
If we let $\bL = {\rm chol}(\frac{1}{P}\bPsi^T \bPsi)$ i.e. $\bL \bL^T = \frac{1}{P} \bPsi^T \bPsi$.  A valid way to sample from a matrix variate normal gives:
\[
\bbeta = s^+(w_a) \bPsi +  s^+(w_b) \bPsi \bzeta \bL^T
\]

\noindent
As mentioned before, we treat the problem in a Bayesian way and approximate the posterior over all the parameters. We use variational inference to transform the problem into an optimization problem.
We use a univariate normal variational approximation on $w_a, w_b, w_c$ with means $\mu_{q_{w_a}}, \mu_{q_{w_b}},\mu_{q_{w_c}}$ and  variance $\sigma_{q_{w_a}}^2, \sigma_{q_{w_b}}^2,\sigma_{q_{w_c}}^2$.  The variational approximation on $\kappa$ is a diagonal covariance multivariate normal with mean given by $\bmu_{q_{\kappa}}$ and covariance given by ${\rm diag}(\bsigma_{q_{\kappa}}^2)$.  Similarly we put a univariate normal variational approximation over each element of $\bzeta$ parameterized so that $\bzeta_{i,j}$ has mean $\mu_{q_{\zeta_{i,j}}}$ and variance $\sigma^2_{q_{\zeta_{i,j}}} $. This gives us $2(P + K^2 + 3)$ parameters to estimate. We denote $Q$ as the Gaussian variational posterior over all of the parameters, and $P$ the prior and maximize : 
\begin{align}\label{lowerbound}
\mathcal{L} = &\E_{Q}[c_n \log {\rm sigmoid}(\lambda_n) + (1-c_n)\log\{ 1-{\rm sigmoid}(\lambda_n)\}] \\ \nonumber
&- \frac{1}{N}{\rm KL}(Q|P),
\end{align}

\noindent
where:
\begin{align*}
\lambda_n & = \bbeta_{a_n} \hat{\bomega}_n + \bkappa_{a_n} \\
& = s^+(w_a) \bPsi_{a_n} \hat{\bomega}_n +    s^+(w_b) \{ (\bL \hat{\bomega}_n )^T \otimes \bPsi_{a_n} \} {\rm vec}( \bzeta)  + \bkappa_{a_n}
\end{align*}
We use the local re-parameterization trick \cite{kingma2015variational} which uses the Affine transform properties of multivariate Gaussian distribution to allow the re-parameterization trick to be employed on lower dimensions.  This results in sampling at lower dimensions and more importantly makes the derivatives of the loss less noisy.
To implement the local re-parameterization trick we draw random samples:
\[
\bepsilon_{w_a} \sim \mathcal{N}(0,1), \hspace{.3cm} \bepsilon_{w_b} \sim \mathcal{N}(0,1), \hspace{.3cm} \bepsilon_{\rm lrt} \sim \mathcal{N}(0,1), \hspace{.3cm} \bepsilon_{\kappa} \sim \mathcal{N}(0,1).
\]

\noindent
with $\bR_n = (\bL \hat{\bomega}_n )^T \otimes \bPsi_{a_n}$, we can get a one dimentional noisy estimate of $\lambda_n$ :
\noindent
\begin{align*}
\hat{\lambda}_n = &  s^+(\mu_{q_{w_a}} + \epsilon_{w_a} \sigma_{q_{w_a}} ) \bPsi_{a_n} \hat{\bomega}_n \\
& + s^+(\mu_{q_{w_b}} + \epsilon_{w_b} \sigma_{q_{w_b}} ) ( \bR_n ~ {\rm vec}( \bmu_{q_\zeta}) +  \| \bR_n^T \odot {\rm vec}(\bsigma_{q_\zeta}) \|_2  \bepsilon_{\rm lrt} ) \\
& + \bmu_{q_{\kappa_a}} + \mu_{q_{w_c}} + \epsilon_{\kappa} \sqrt{\bsigma_{q_{\kappa_a}}^2 + \sigma^2_{q_{w_c}}}.
\end{align*}

\noindent
where $\|\cdot\|_2$ denotes the $L_2$ norm and $\odot$ element wise multiplication.  We can optimize a noisy version of our objective :
\vspace{-1mm}
\begin{align}\label{mclowerbound}
\hat{\mathcal{L}}_n =& c_n \log {\rm sigmoid}(\hat{\lambda}_n) + (1-c_n)\log\{ 1-{\rm sigmoid}(\hat{\lambda}_n)\} \\ \nonumber &- \frac{1}{N}{\rm KL}(Q|P).
\end{align}

We call the solution of this optimization problem \textbf{BLOB-NQ} as we considered a Normal approximation for the posterior on $\zeta$.\\

\noindent
An alternative approach is to use a matrix variate normal distribution as the variational approximation of $\bzeta$ with mean matrix $\bmu_{q_\zeta}$ and the two covariance matrices given by: ${\rm diag}(\bsigma^2_{q_{\zeta_1}})$ and  ${\rm diag} (\bsigma^2_{q_{\zeta_2} })$.  This reduces the number of variational parameters used for representing the variance of the variational posterior. We thus need to estimate $2(P + 3) + K^2 + 2K$ which is less then the previous approximation for $K \ge 2$.  To apply the local re-parameterization trick let:
\vspace{-1mm}
\[
std_n = \sqrt{(\bsigma_{q_{\zeta_1 }}^2 \cdot \bPsi_{a_n}^2)(\bsigma_{q_{\zeta_2}}^2 \cdot (\bL^T \hat{\bomega}_n)^2})
\]
\begin{align*}
\hat{\phi}_n = &  s^+(\mu_{q_{w_a}} + \epsilon_{w_a} \sigma_{q_{w_a}} ) \bPsi_{a_n} \hat{\bomega}_n \\
& + s^+(\mu_{q_{w_b}} + \epsilon_{w_b} \sigma_{q_{w_b}} ) \{ \bPsi_{a_n}\bmu_{q_\zeta} \bL^T \hat{\bomega}_n +  std_n \bepsilon_{\rm lrt} \} \\
& + \bmu_{q_{\kappa_a}} + \mu_{q_{w_c}} + \epsilon_{\kappa} \sqrt{\bsigma_{q_{\kappa_a}}^2 + \sigma^2_{q_{w_c}}}.
\end{align*}

\noindent
A noisy estimate of the lower bound can then be computed by substituting $\hat{\phi}_n$ into Equation \eqref{mclowerbound}. We call its solution \textbf{BLOB-MNQ} as we use a Matrix Normal variational posterior.


In both approximations and when the objective is at its maximum, we can take a point estimate of the bandit embeddings:

\[
\hat{\bbeta} = s^+(\mu_{q_{w_a}}) \bPsi + s^+(\mu_{q_{w_b}}) \bPsi \bmu_{q_\zeta} \bL^T.
\]

\noindent
The bandit embedding can be interpreted as a weighted sum of the organic embedding and the organic embedding multiplied by a $K \times K$ matrix that can adjust the bandit embeddings based on the bandit signal.

\section{Results}
\subsection{Organic Evaluation}
We demonstrate that our method produces useful user representations on next item prediction using the RecoGym simulation environment \cite{rohde2018recogym}. RecoGym is a framework for simulating a recommender system and enables the simulation of A/B tests although here we simply use it to create organic sequences of item views and test the organic model's ability to do next item prediction.  We split both the datasets into train and test so that sessions reside entirely in one of the two groups.  We fit the model to the training set, we then evaluate by providing the model $v_1,..v_{{T_u}-1}$ events and testing the model's ability to predict $v_{T_{u}}$.

The organic model was implemented using the PyTorch automatic differentiation package in Python \cite{paszke2017automatic} and trained using Stochastic Gradient Descent (SGD), specifically the RMSProp variant. We set the learning rate to 0.001 and tune the other hyper-parameters, including L2 regularization, for each dataset based upon a validation set\footnote{Source code: \url{https://github.com/criteo-research/blob}).  The RecoGym simulator allows reproducible results for all recommendation algorithms and policies. }.

The various models are evaluated using recall at K (RC@K) and truncated discounted cumulative gain at K (DCG@K), which are defined below.

Let $r_k$ be the $k$th highest value of $p(\bomega_{v_{T_{u}}}|v_1,..v_{{T_u}-1})$. For all results presented in this paper, we set K to 5.

\begin{align*}
  {\rm RC@K} =\begin{cases}
    1, & \text{if $v_{T_u} \in \{r_1,...,r_K\}$}.\\
    0, & \text{otherwise}.
  \end{cases}
\end{align*}

\begin{align*}
  {\rm DCG@K} =
    \sum_i \frac{2^{r_i \boldsymbol{1}\{v_{T_u}\in \{r_1,...,r_K\} \} }-1 }{\log i+1}.
\end{align*}

\noindent
We compute the average of these quantities over all sessions in the test set.


We consider two alternative methods for training the model:

\begin{itemize}
  \item \textbf{Bouch/AE} - A linear variational auto-encoder using the Bouchard bound (see the supplementary material).
  \item \textbf{RT/AE} - A deep auto-encoder again using the re-parameterization trick. The deep auto-encoder consists of mapping an input of size P to three linear rectifier layers of K units each.
\end{itemize}

When we update the posterior over a user's  latent variable representation at test time, we assess both using the auto-encoder denoted AE and using the 100 iterations of the EM algorithm denoted EM in the results.

When we compute next item predictions we consider both using a 100 sample Monte Carlo approximation denoted MC and just taking the mean as a point estimate denoted mean it uses only $\bmu_{q_\omega}$ (and correspondingly ignores $\bSigma_q$).

To demonstrate the effectiveness of our approach, we present results from the following baseline approaches:

\textbf{Popularity}:  Item popularity provides no personalization, but is nonetheless a strong baselines for certain recommendation tasks.

\textbf{Item KNN}: Item K Nearest Neighbors (KNN) involves computing the correlation matrix of the sample data adding the identity to prevent division by zero and then using these correlations as recommendations based on a user's most recent historical item.  The limitations of this technique is that it ignores item popularity and multiple items in the user's history, but despite these limitations it is often a strong baseline.

\textbf{Recurrent Neural Network}: For this baseline, we make use of a recurrent neural network to learn a user representation by predicting the next item in the session. The model architecture we employ is similar to that of \cite{hidasi2018recurrent}, in that we feed the output from an embedding layer into a Gated Recurrent Unit (GRU) \cite{cho2014learning} with 64 hidden units to learn the temporal dynamics of the user's session. The output from the GRU is then passed through a final softmax layer which gives the probability of the next item in the sequence. The network is trained to minimize the categorical cross-entropy over the training sessions via RMSProp.

\begin{table}
\begin{tabular}{lllrr}
  \toprule
  \textbf{Train}  & \textbf{Online}  & \textbf{Online}  &   \textbf{RC@5} &  \textbf{DCG@5} \\
  \textbf{Algorithm} &  \textbf{Latent} & \textbf{Next Item} &    &   \\
\midrule
\midrule
Pop &          &        &  0.020 &   0.016 \\
ItemKNN &          &        &  0.020 &   0.024 \\
RNN &          &        &  0.035 &   0.033 \\
\midrule
Bouch/AE &          AE &        MC &  0.082 &   0.128 \\
Bouch/AE &          AE &      mean &  0.082 &   0.079 \\
Bouch/AE &          EM &        MC &  \textbf{0.117} &   0.128 \\
Bouch/AE &          EM &      mean &  \textbf{0.117} &   \textbf{0.130}  \\
RT/AE &          AE &        MC &  0.090 &   0.105 \\
RT/AE &          AE &      mean &  0.080 &   0.068 \\
RT/AE &          EM &        MC &  0.090 &   0.105 \\
RT/AE &          EM &      mean &  0.090 &   0.106 \\
\bottomrule 
  \end{tabular}
  \caption{Results on the testset of RecoGym dataset with 2000 products. For both metrics, a higher value is better.}
  \label{recogym2000}  
 \end{table}

For our organic experiment we use the RecoGym simulator with 2000 products and $\sigma_\omega=0$, i.e. a static user state, we generate a training set of 100 sessions and a test set of 100 sessions, this results in 21852 and 19533 events for train and test respectively.  The BLO models were all trained using 15000 epochs using the RMSProp algorithm, the embedding size was set to 10.  The RNN was trained with K=200 for 5000 epochs (it performed slightly worse with a training run of 25000). The results are shown in Table 2.  BLO is much better than the baselines at standard organic recommender systems metrics.  However if being able to build an adequate model of organic behaviour is sufficient for building a recommender system depends on if the organic behaviour is aligned with bandit behaviour.  This requires using RecoGym for its intended purpose simulating A/B tests and varying the agreement between the organic behavior and bandit behavior using the provided flips parameter.


\subsection{The Complete Model - Organic and Bandit}
\subsubsection{Experimental Setup}
Unfortunately no real world dataset exhibits the required properties (both organic and bandit behavior) moreover no real world dataset including counterfactual datasets allow us to evaluate the quality of a recommender systems recommendations reliably.  For this reason for the complete dataset we do our evaluations completely in the RecoGym simulator.  A strong advantage of the simulation environment is that not only can we compute offline organic metrics but we can also simulate A/B tests.

Another advantage of the RecoGym simulator that simulates both organic and bandit behaviour is that algorithms from the traditional organic part of recommender systems research and bandit algorithms can be compared side by side.  We consider traditional organic algorithms like ItemKNN \cite{davidson2010youtube} along side our organic Bayesian Latent Organic model (BLO) and sophisticated deep learning approaches such as the MultiVAE \cite{Liang2018}.  In the case of bandit algorithms we can test value based logistic regression as well as the policy based contextual bandit.  In order to apply any bandit algorithm we need to perform feature engineering in order to transform the history consisting of item views into a vector of history.  For the logistic regression we elect to make a $P$ dimensional feature vector crossed with the action also of size $P$ giving $P^2$ features.  Similarly the contextual bandit is a linear model that maps the $P$ dimensional vector of historical counts to a $P$ dimensional action space.

We are interested to see how the recommender system responds to different logging policies, we therefore test it using a good logging policy based on the session popularity.  That is the probability $1-\epsilon$ is shared proportionally to the items in a users history we use considerable exploration ($\epsilon=0.3$).  We are interested in the (common) case where we have plentiful organic data so we set RecoGym to have 20000 organic sessions.  Finally we are interested in situations where the next item prediction is an optimal recommendation and cases where the organic signal alone is misleading to recommendation quality.  This connection between the organic and the bandit signal is controlled with the \emph{flips} parameter in RecoGym.  The flips parameter permutes the behavior of two actions.

A unique feature of RecoGym is that we are able to simulate both organic and bandit feedback, this means we are able to compare algorithms that operate on the bandit signal (both policy and value based) with algorithms that operate on the organic signal.  We consider the following baselines:

\textbf{Logistic regression (bandit, value)}:
Perhaps the simplest way to process a bandit signal.  We regress the reward on features derived from the users history and the recommended action.  In order to deliver the recommendation we predict the reward for every action and select the highest.

\textbf{Contextual bandit (bandit, policy)}:
The contextual bandit is a policy based method that maps a context to a recommendation in one-of-n coding a vector of length $P$.  The algorithm is trained using counterfactual risk minimization using the IPS score logged by RecoGym without any clipping or variance penalty.

\textbf{Session ItemKNN (organic)}:  This organic algorithm operates by determining for each session if an item was present or absent, from this dataset a correlation matrix is computed.  At recommendation is delivered by computing the average correlations for each item in history as a single vector and then taking the maximum.  We take the whole session into account rather than the most recent item (unlike most recent ItemKNN used above).

\textbf{Multi-VAE (organic)}:  A state of the art deep learning recommendation algorithm similar to the organic portion of the model presented here except the model is non-linear and uses some non-standard heuristics such as ``beta-annealing''.

\textbf{BLO (organic)}:  
The organic portion of the model developed here.  We set the embedding size to be K=20 and use a linear variational auto-encoder.  This is implemented in PyTorch.  A learning rate of 0.0001 is used with 1000 epochs and an embedding size of $K=20$.

\textbf{BLOB (organic and bandit combined)}:  
The complete model developed here.  We use priors: 
$w_a \sim \mathcal{N}(-1,1^2)$
$w_b \sim \mathcal{N}(-6,1^2)$
$w_c \sim \mathcal{N}(-4.5,10^2)$
$\kappa \sim \mathcal{N}(w_c,0.01^2 \bI)$.  We consider both the normal variational approximation NQ and the matrix normal variational approximation MNQ.  The bandit layer is implemented using TensorFlow with a learning rate of 0.001 and 800 epochs for the P=100 and 1200 epochs for the P=1000, with a batch size of 1024 and using the RMSprop training algorithm.

\textbf{Random}:  
The actions are recommended randomly. A weak baseline but useful to calibrate performance.
\subsubsection{Experimental Results}
The first experiment considers the catalog size to be P=100, the number of user sessions to be 1000, the simulated A/B test is done over 4000 users and the logging policy being session popularity with epsilon greedy exploration (epsilon=0.3).    
This means that the bandit signal will resemble that found in real systems with a strong signal around some actions favoured by the previous version of the recommender system (session popularity policy - a decent baseline) and a weak signal over much of the remaining action space.  Results are shown in Table 3.

In the Flips=0 scenario RecoGym is configured so that next item prediction based on organic data is a perfect proxy  for delivering good recommendations.  As a consequence all the organic based methods do well including the BLO (organic), both our methods that combine organic and bandit BLOB-NQ and BLOB-MNQ and the Multi-VAE baseline, the Session ItemKNN baseline while organic does not perform well.

When the Flips=50 scenario RecoGym internally permutes 50 actions behavior this means that next item prediction is now a poor proxy of recommendation performance.  We see this as all purely organic based agents now perform poorly indeed the connection between organic and bandit is reduced to the point that Session ItemKNN, the Multi VAE and BLO all perform worse than random.  It is in this case that the value of our BLOB model is demonstrated as both BLOB NQ and BLOB MNQ perform strongly. 

For the purely signal bandit algorithms the value based Log Reg and the policy based CB perform similarly to each other and with Flips=0 and Flips=50.  They perform a little better than random (except for CB Flips=50) demonstrating that there is some usable signal in the bandit feed back but are far from state of the art especially in the Flips=0 case where ignoring the organic signal profoundly limits recommendation quality.  In the Flips=50 case the pure bandit approaches outperform the purely organic algorithms but the combined approach performs significantly better giving a click through rate of 1.57\% for the BLOB NQ compared to 1.21\% for the logistic regression.

Importantly the BLOB NQ and BLOB MNQ outperform or equal the other methods in the Flips=0 setting and outperform the other methods in the Flips=50 setting.

\begin{table}[ht]
\centering
\caption{Simulated A/B test results on the RecoGym simulator using: P=100, U=1000, organic only sessions=20 000. }
\label{results1000}
\begin{tabular}{llrr}
  \hline
  Agent & Type & CTR (\%)  & CTR (\%) \\
   & & Flips=0       & Flips=50\\
  \hline
  Log Reg & (bandit) & 1.37 & 1.21 \\
    CB & (bandit) & 1.37 & 1.09 \\
    \hline
    ItemKNN & (organic) & 1.39 & 0.92 \\
    MultiVAE & (organic) & \textbf{2.43} & 0.76 \\
    BLO & (organic) & \textbf{2.42} & 0.76 \\
    \hline
    BLOB-NQ & (combined) & \textbf{2.42} & \textbf{1.57} \\
    BLOB-MNQ & (combined)  & \textbf{2.40} & \textbf{1.56} \\
    \hline
    Random & & 1.09 & 1.11 \\
   \hline
\end{tabular}
\end{table}

The second experiment considers the same setup but with $P=1000$, we also increase the number of epochs on the bandit component of the model to 1200.  Results are shown in Table 4.  

Again we see that the methods that use the organic data either the purely organic or the combined BLOB methods we propose perform work well when Flips=0, but when Flips=500 the purely organic methods fall in performance to little above random yet the combined methods BLOB-MNQ and BLOB-NQ continue to perform well beating all other baselines.

The policy based contextual bandit shows a small improvement over the value based logistic regression in the Flips=0 case although this advantage vanishes when Flips=500, this is may be due to the fact that the contextual bandit ``clings'' to the logging policy and the session popularity logging policy is better in the case where Flips=0.

\begin{table}[ht]
\centering
\caption{Simulated A/B test results on the RecoGym simulator using: P=1000, U=1000, organic only sessions=20 000. }
\begin{tabular}{llrr}
  \hline
  Agent & Type & CTR (\%)  & CTR (\%) \\
   & & Flips=0       & Flips=500\\
  \hline
  Log Reg & (bandit) & 1.26 & 1.30 \\
    CB & (bandit) & 1.38 & 1.29 \\
    \hline
    ItemKNN & (organic) & 1.39 & 0.87 \\
    MultiVAE & (organic) & \textbf{2.43} & 1.15 \\
    BLO & (organic) & \textbf{2.42} & 1.13 \\
    \hline
    BLOB-NQ& (combined)  & \textbf{2.40} & 1.51 \\
    BLOB-MNQ& (combined)  & \textbf{2.39} & \textbf{1.62} \\
    \hline
    Random & & 1.13 & 1.12 \\
   \hline
\end{tabular}
\label{sim1000prod}
\end{table}

\section{Conclusion}
We focus on a particular recommendation task, one where a user profile is defined by a history of items in a catalog and the recommendation task is to recommend items from the same catalog. Our model is able to learn both from the organic signal and the bandit signal jointly beating baselines in a range of settings by exploiting the three fundamental distances of recommendation action-history, action-action and history-history.

We use computational techniques which allow allow large scale Bayesian inference suitable for Recommendation with large catalogs. The local re-parameterization trick was particularly valuable in reducing the variance in our optimisation problem.

BLOB is able to perform well both in situations where next item prediction is a good proxy for recommendations and situations where it is poor.  Meeting the performance of pure organic algorithms in settings where the organic signal is sufficient and exceeding all baselines organic and bandit (policy) and bandit (value).  This strongly validates the value of Bayesian methods to infer in the cases of a signal of varying strength and their practical value thanks to modern developments in Bayesian deep learning.

There are many possible extension to this work, one is to produce end to end training i.e. training both the organic and bandit component simultaneously.  To apply this approach would require a more complicated training procedure.  We also expect there are other useful ways to combine organic and bandit signal, perhaps based on models that avoid the softmax and sigmoid transform such as LDA for the organic and using the approach out lined in \cite{lumbreras2018bayesian} for the Bandit.  Avoiding softmax and sigmoid transforms has both computational advantages and can increase interpretability.


\bibliographystyle{ACM-Reference-Format}
\bibliography{bibliography}


\begin{thebibliography}{00}


\ifx \showCODEN    \undefined \def \showCODEN     #1{\unskip}     \fi
\ifx \showDOI      \undefined \def \showDOI       #1{{\tt DOI:}\penalty0{#1}\ }
  \fi
\ifx \showISBNx    \undefined \def \showISBNx     #1{\unskip}     \fi
\ifx \showISBNxiii \undefined \def \showISBNxiii  #1{\unskip}     \fi
\ifx \showISSN     \undefined \def \showISSN      #1{\unskip}     \fi
\ifx \showLCCN     \undefined \def \showLCCN      #1{\unskip}     \fi
\ifx \shownote     \undefined \def \shownote      #1{#1}          \fi
\ifx \showarticletitle \undefined \def \showarticletitle #1{#1}   \fi
\ifx \showURL      \undefined \def \showURL       #1{#1}          \fi
\providecommand\bibfield[2]{#2}
\providecommand\bibinfo[2]{#2}
\providecommand\natexlab[1]{#1}
\providecommand\showeprint[2][]{arXiv:#2}

\bibitem[\protect\citeauthoryear{Adomavicius and Tuzhilin}{Adomavicius and
  Tuzhilin}{2005}]%
        {adomavicius2005toward}
\bibfield{author}{\bibinfo{person}{Gediminas Adomavicius} {and}
  \bibinfo{person}{Alexander Tuzhilin}.} \bibinfo{year}{2005}\natexlab{}.
\newblock \showarticletitle{Toward the next generation of recommender systems:
  A survey of the state-of-the-art and possible extensions}.
\newblock \bibinfo{journal}{{\em IEEE transactions on knowledge and data
  engineering\/}} \bibinfo{volume}{17}, \bibinfo{number}{6}
  (\bibinfo{year}{2005}), \bibinfo{pages}{734--749}.
\newblock


\bibitem[\protect\citeauthoryear{Berger, Wolpert, Bayarri, DeGroot, Hill, Lane,
  and LeCam}{Berger et~al\mbox{.}}{1988}]%
        {berger1988likelihood}
\bibfield{author}{\bibinfo{person}{James~O Berger}, \bibinfo{person}{Robert~L
  Wolpert}, \bibinfo{person}{MJ Bayarri}, \bibinfo{person}{MH DeGroot},
  \bibinfo{person}{Bruce~M Hill}, \bibinfo{person}{David~A Lane}, {and}
  \bibinfo{person}{Lucien LeCam}.} \bibinfo{year}{1988}\natexlab{}.
\newblock \showarticletitle{The likelihood principle}.
\newblock \bibinfo{journal}{{\em Lecture notes-Monograph series\/}}
  \bibinfo{volume}{6} (\bibinfo{year}{1988}), \bibinfo{pages}{iii--199}.
\newblock


\bibitem[\protect\citeauthoryear{Beygelzimer and Langford}{Beygelzimer and
  Langford}{2009}]%
        {beygelzimer2009offset}
\bibfield{author}{\bibinfo{person}{Alina Beygelzimer} {and}
  \bibinfo{person}{John Langford}.} \bibinfo{year}{2009}\natexlab{}.
\newblock \showarticletitle{The offset tree for learning with partial labels}.
  In \bibinfo{booktitle}{{\em Proceedings of the 15th ACM SIGKDD international
  conference on Knowledge discovery and data mining}}. ACM,
  \bibinfo{pages}{129--138}.
\newblock


\bibitem[\protect\citeauthoryear{Blei and Lafferty}{Blei and Lafferty}{2005}]%
        {lafferty2006correlatedold}
\bibfield{author}{\bibinfo{person}{David~M. Blei} {and}
  \bibinfo{person}{John~D. Lafferty}.} \bibinfo{year}{2005}\natexlab{}.
\newblock \showarticletitle{Correlated Topic Models}. In
  \bibinfo{booktitle}{{\em Advances in Neural Information Processing Systems 18
  [Neural Information Processing Systems, {NIPS} 2005, December 5-8, 2005,
  Vancouver, British Columbia, Canada]}}. \bibinfo{pages}{147--154}.
\newblock
\showURL{%
\url{http://papers.nips.cc/paper/2906-correlated-topic-models}}


\bibitem[\protect\citeauthoryear{Blei, Ng, and Jordan}{Blei
  et~al\mbox{.}}{2003}]%
        {blei2003latent}
\bibfield{author}{\bibinfo{person}{David~M Blei}, \bibinfo{person}{Andrew~Y
  Ng}, {and} \bibinfo{person}{Michael~I Jordan}.}
  \bibinfo{year}{2003}\natexlab{}.
\newblock \showarticletitle{Latent dirichlet allocation}.
\newblock \bibinfo{journal}{{\em Journal of machine Learning research\/}}
  \bibinfo{volume}{3}, \bibinfo{number}{Jan} (\bibinfo{year}{2003}),
  \bibinfo{pages}{993--1022}.
\newblock


\bibitem[\protect\citeauthoryear{Bottou, Peters, Qui{\~n}onero-Candela,
  Charles, Chickering, Portugaly, Ray, Simard, and Snelson}{Bottou
  et~al\mbox{.}}{2013}]%
        {Bottou2013}
\bibfield{author}{\bibinfo{person}{L. Bottou}, \bibinfo{person}{J. Peters},
  \bibinfo{person}{J. Qui{\~n}onero-Candela}, \bibinfo{person}{D. Charles},
  \bibinfo{person}{D. Chickering}, \bibinfo{person}{E. Portugaly},
  \bibinfo{person}{D. Ray}, \bibinfo{person}{P. Simard}, {and}
  \bibinfo{person}{E. Snelson}.} \bibinfo{year}{2013}\natexlab{}.
\newblock \showarticletitle{Counterfactual reasoning and learning systems: The
  example of computational advertising}.
\newblock \bibinfo{journal}{{\em The Journal of Machine Learning Research\/}}
  \bibinfo{volume}{14}, \bibinfo{number}{1} (\bibinfo{year}{2013}),
  \bibinfo{pages}{3207--3260}.
\newblock


\bibitem[\protect\citeauthoryear{Bouchard}{Bouchard}{2007}]%
        {bouchard2007efficient}
\bibfield{author}{\bibinfo{person}{Guillaume Bouchard}.}
  \bibinfo{year}{2007}\natexlab{}.
\newblock \bibinfo{title}{Efficient bounds for the softmax function,
  applications to inference in hybrid models}.
\newblock   (\bibinfo{year}{2007}).
\newblock


\bibitem[\protect\citeauthoryear{Capp{\'e} and Moulines}{Capp{\'e} and
  Moulines}{2009}]%
        {cappe2009line}
\bibfield{author}{\bibinfo{person}{Olivier Capp{\'e}} {and}
  \bibinfo{person}{Eric Moulines}.} \bibinfo{year}{2009}\natexlab{}.
\newblock \showarticletitle{On-line expectation--maximization algorithm for
  latent data models}.
\newblock \bibinfo{journal}{{\em Journal of the Royal Statistical Society:
  Series B (Statistical Methodology)\/}} \bibinfo{volume}{71},
  \bibinfo{number}{3} (\bibinfo{year}{2009}), \bibinfo{pages}{593--613}.
\newblock


\bibitem[\protect\citeauthoryear{Cho, van Merrienboer, G{\"{u}}l{\c{c}}ehre,
  Bahdanau, Bougares, Schwenk, and Bengio}{Cho et~al\mbox{.}}{2014}]%
        {cho2014learning}
\bibfield{author}{\bibinfo{person}{Kyunghyun Cho}, \bibinfo{person}{Bart van
  Merrienboer}, \bibinfo{person}{{\c{C}}aglar G{\"{u}}l{\c{c}}ehre},
  \bibinfo{person}{Dzmitry Bahdanau}, \bibinfo{person}{Fethi Bougares},
  \bibinfo{person}{Holger Schwenk}, {and} \bibinfo{person}{Yoshua Bengio}.}
  \bibinfo{year}{2014}\natexlab{}.
\newblock \showarticletitle{Learning Phrase Representations using {RNN}
  Encoder-Decoder for Statistical Machine Translation}. In
  \bibinfo{booktitle}{{\em Proceedings of the 2014 Conference on Empirical
  Methods in Natural Language Processing, {EMNLP} 2014, October 25-29, 2014,
  Doha, Qatar, {A} meeting of SIGDAT, a Special Interest Group of the {ACL}}}.
  \bibinfo{publisher}{{ACL}}, \bibinfo{pages}{1724--1734}.
\newblock
\showURL{%
\url{http://aclweb.org/anthology/D/D14/D14-1179.pdf}}


\bibitem[\protect\citeauthoryear{Davidson, Liebald, Liu, Nandy, Van~Vleet,
  Gargi, Gupta, He, Lambert, Livingston, et~al\mbox{.}}{Davidson
  et~al\mbox{.}}{2010}]%
        {davidson2010youtube}
\bibfield{author}{\bibinfo{person}{James Davidson}, \bibinfo{person}{Benjamin
  Liebald}, \bibinfo{person}{Junning Liu}, \bibinfo{person}{Palash Nandy},
  \bibinfo{person}{Taylor Van~Vleet}, \bibinfo{person}{Ullas Gargi},
  \bibinfo{person}{Sujoy Gupta}, \bibinfo{person}{Yu He}, \bibinfo{person}{Mike
  Lambert}, \bibinfo{person}{Blake Livingston}, {and}
  \bibinfo{person}{others}.} \bibinfo{year}{2010}\natexlab{}.
\newblock \showarticletitle{The YouTube video recommendation system}. In
  \bibinfo{booktitle}{{\em Proceedings of the fourth ACM conference on
  Recommender systems}}. \bibinfo{pages}{293--296}.
\newblock


\bibitem[\protect\citeauthoryear{Gionis, Indyk, Motwani, et~al\mbox{.}}{Gionis
  et~al\mbox{.}}{1999}]%
        {gionis1999similarity}
\bibfield{author}{\bibinfo{person}{Aristides Gionis}, \bibinfo{person}{Piotr
  Indyk}, \bibinfo{person}{Rajeev Motwani}, {and} \bibinfo{person}{others}.}
  \bibinfo{year}{1999}\natexlab{}.
\newblock \showarticletitle{Similarity search in high dimensions via hashing}.
  In \bibinfo{booktitle}{{\em Vldb}}, Vol.~\bibinfo{volume}{99}.
  \bibinfo{pages}{518--529}.
\newblock


\bibitem[\protect\citeauthoryear{Hernan and Robins}{Hernan and Robins}{2010}]%
        {hernan2010causal}
\bibfield{author}{\bibinfo{person}{Miguel~A Hernan} {and}
  \bibinfo{person}{James~M Robins}.} \bibinfo{year}{2010}\natexlab{}.
\newblock \bibinfo{title}{Causal inference}.
\newblock   (\bibinfo{year}{2010}).
\newblock


\bibitem[\protect\citeauthoryear{Hidasi and Karatzoglou}{Hidasi and
  Karatzoglou}{2018}]%
        {hidasi2018recurrent}
\bibfield{author}{\bibinfo{person}{Bal{\'{a}}zs Hidasi} {and}
  \bibinfo{person}{Alexandros Karatzoglou}.} \bibinfo{year}{2018}\natexlab{}.
\newblock \showarticletitle{Recurrent Neural Networks with Top-k Gains for
  Session-based Recommendations}. In \bibinfo{booktitle}{{\em Proceedings of
  the 27th {ACM} International Conference on Information and Knowledge
  Management, {CIKM} 2018, Torino, Italy, October 22-26, 2018}},
  \bibfield{editor}{\bibinfo{person}{Alfredo Cuzzocrea}, \bibinfo{person}{James
  Allan}, \bibinfo{person}{Norman~W. Paton}, \bibinfo{person}{Divesh
  Srivastava}, \bibinfo{person}{Rakesh Agrawal}, \bibinfo{person}{Andrei~Z.
  Broder}, \bibinfo{person}{Mohammed~J. Zaki}, \bibinfo{person}{K.~Sel{\c{c}}uk
  Candan}, \bibinfo{person}{Alexandros Labrinidis}, \bibinfo{person}{Assaf
  Schuster}, {and} \bibinfo{person}{Haixun Wang}} (Eds.).
  \bibinfo{publisher}{{ACM}}, \bibinfo{pages}{843--852}.
\newblock
\showISBNx{978-1-4503-6014-2}
\showDOI{%
\url{http://dx.doi.org/10.1145/3269206.3271761}}


\bibitem[\protect\citeauthoryear{Jaakkola and Jordan}{Jaakkola and
  Jordan}{1997}]%
        {jaakkola1997variational}
\bibfield{author}{\bibinfo{person}{Tommi Jaakkola} {and}
  \bibinfo{person}{Michael Jordan}.} \bibinfo{year}{1997}\natexlab{}.
\newblock \showarticletitle{A variational approach to Bayesian logistic
  regression models and their extensions}. In \bibinfo{booktitle}{{\em Sixth
  International Workshop on Artificial Intelligence and Statistics}},
  Vol.~\bibinfo{volume}{82}. \bibinfo{pages}{4}.
\newblock


\bibitem[\protect\citeauthoryear{Jeunen, Rohde, Vasile, and Bompaire}{Jeunen
  et~al\mbox{.}}{2020}]%
        {Jeunen2020}
\bibfield{author}{\bibinfo{person}{O. Jeunen}, \bibinfo{person}{D. Rohde},
  \bibinfo{person}{F. Vasile}, {and} \bibinfo{person}{M. Bompaire}.}
  \bibinfo{year}{2020}\natexlab{}.
\newblock \showarticletitle{Joint Policy-Value Learning for Recommendation}. In
  \bibinfo{booktitle}{{\em Proc. of the 26th ACM SIGKDD International
  Conference on Knowledge Discovery \& Data Mining}} {\em (\bibinfo{series}{KDD
  '20})}.
\newblock


\bibitem[\protect\citeauthoryear{Johansson, Shalit, and Sontag}{Johansson
  et~al\mbox{.}}{2016}]%
        {johansson2016learning}
\bibfield{author}{\bibinfo{person}{Fredrik Johansson}, \bibinfo{person}{Uri
  Shalit}, {and} \bibinfo{person}{David Sontag}.}
  \bibinfo{year}{2016}\natexlab{}.
\newblock \showarticletitle{Learning representations for counterfactual
  inference}. In \bibinfo{booktitle}{{\em International Conference on Machine
  Learning}}. \bibinfo{pages}{3020--3029}.
\newblock


\bibitem[\protect\citeauthoryear{Kingma, Salimans, and Welling}{Kingma
  et~al\mbox{.}}{2015}]%
        {kingma2015variational}
\bibfield{author}{\bibinfo{person}{Diederik~P Kingma}, \bibinfo{person}{Tim
  Salimans}, {and} \bibinfo{person}{Max Welling}.}
  \bibinfo{year}{2015}\natexlab{}.
\newblock \showarticletitle{Variational dropout and the local
  reparameterization trick}. In \bibinfo{booktitle}{{\em Advances in Neural
  Information Processing Systems}}. \bibinfo{pages}{2575--2583}.
\newblock


\bibitem[\protect\citeauthoryear{Kingma and Welling}{Kingma and
  Welling}{2014}]%
        {kingma2013auto}
\bibfield{author}{\bibinfo{person}{Diederik~P. Kingma} {and}
  \bibinfo{person}{Max Welling}.} \bibinfo{year}{2014}\natexlab{}.
\newblock \showarticletitle{Auto-Encoding Variational Bayes}. In
  \bibinfo{booktitle}{{\em 2nd International Conference on Learning
  Representations, {ICLR} 2014, Banff, AB, Canada, April 14-16, 2014,
  Conference Track Proceedings}}, \bibfield{editor}{\bibinfo{person}{Yoshua
  Bengio} {and} \bibinfo{person}{Yann LeCun}} (Eds.).
\newblock
\showURL{%
\url{https://openreview.net/group?id=ICLR.cc/2014}}


\bibitem[\protect\citeauthoryear{Knowles and Minka}{Knowles and Minka}{2011}]%
        {knowles2011non}
\bibfield{author}{\bibinfo{person}{David~A. Knowles} {and} \bibinfo{person}{Tom
  Minka}.} \bibinfo{year}{2011}\natexlab{}.
\newblock \showarticletitle{Non-conjugate Variational Message Passing for
  Multinomial and Binary Regression}.
\newblock In \bibinfo{booktitle}{{\em Advances in Neural Information Processing
  Systems 24}}, \bibfield{editor}{\bibinfo{person}{J.~Shawe-Taylor},
  \bibinfo{person}{R.~S. Zemel}, \bibinfo{person}{P.~L. Bartlett},
  \bibinfo{person}{F.~Pereira}, {and} \bibinfo{person}{K.~Q. Weinberger}}
  (Eds.). \bibinfo{publisher}{Curran Associates, Inc.},
  \bibinfo{pages}{1701--1709}.
\newblock


\bibitem[\protect\citeauthoryear{Koren and Bell}{Koren and Bell}{2015}]%
        {koren2015advances}
\bibfield{author}{\bibinfo{person}{Yehuda Koren} {and} \bibinfo{person}{Robert
  Bell}.} \bibinfo{year}{2015}\natexlab{}.
\newblock \showarticletitle{Advances in collaborative filtering}.
\newblock In \bibinfo{booktitle}{{\em Recommender systems handbook}}.
  \bibinfo{publisher}{Springer}, \bibinfo{pages}{77--118}.
\newblock


\bibitem[\protect\citeauthoryear{Lafferty and Blei}{Lafferty and Blei}{2006}]%
        {lafferty2006correlated}
\bibfield{author}{\bibinfo{person}{John~D. Lafferty} {and}
  \bibinfo{person}{David~M. Blei}.} \bibinfo{year}{2006}\natexlab{}.
\newblock \showarticletitle{Correlated Topic Models}. In
  \bibinfo{booktitle}{{\em Advances in Neural Information Processing Systems
  18}}, \bibfield{editor}{\bibinfo{person}{Y.~Weiss},
  \bibinfo{person}{B.~Sch\"{o}lkopf}, {and} \bibinfo{person}{J.~C. Platt}}
  (Eds.). \bibinfo{publisher}{MIT Press}, \bibinfo{pages}{147--154}.
\newblock
\showURL{%
\url{http://papers.nips.cc/paper/2906-correlated-topic-models.pdf}}


\bibitem[\protect\citeauthoryear{Lattimore and Szepesv{\'a}ri}{Lattimore and
  Szepesv{\'a}ri}{2018}]%
        {lattimore2018bandit}
\bibfield{author}{\bibinfo{person}{Tor Lattimore} {and} \bibinfo{person}{Csaba
  Szepesv{\'a}ri}.} \bibinfo{year}{2018}\natexlab{}.
\newblock \showarticletitle{Bandit algorithms}.
\newblock \bibinfo{journal}{{\em preprint\/}} (\bibinfo{year}{2018}),
  \bibinfo{pages}{28}.
\newblock


\bibitem[\protect\citeauthoryear{Liang, Krishnan, Hoffman, and Jebara}{Liang
  et~al\mbox{.}}{2018a}]%
        {liang2018variational}
\bibfield{author}{\bibinfo{person}{Dawen Liang}, \bibinfo{person}{Rahul~G.
  Krishnan}, \bibinfo{person}{Matthew~D. Hoffman}, {and} \bibinfo{person}{Tony
  Jebara}.} \bibinfo{year}{2018}\natexlab{a}.
\newblock \showarticletitle{Variational Autoencoders for Collaborative
  Filtering}. In \bibinfo{booktitle}{{\em Proceedings of the 2018 World Wide
  Web Conference on World Wide Web, {WWW} 2018, Lyon, France, April 23-27,
  2018}}, \bibfield{editor}{\bibinfo{person}{Pierre{-}Antoine Champin},
  \bibinfo{person}{Fabien~L. Gandon}, \bibinfo{person}{Mounia Lalmas}, {and}
  \bibinfo{person}{Panagiotis~G. Ipeirotis}} (Eds.).
  \bibinfo{publisher}{{ACM}}, \bibinfo{pages}{689--698}.
\newblock
\showDOI{%
\url{http://dx.doi.org/10.1145/3178876.3186150}}


\bibitem[\protect\citeauthoryear{Liang, Krishnan, Hoffman, and Jebara}{Liang
  et~al\mbox{.}}{2018b}]%
        {Liang2018}
\bibfield{author}{\bibinfo{person}{D. Liang}, \bibinfo{person}{R.~G. Krishnan},
  \bibinfo{person}{M.~D Hoffman}, {and} \bibinfo{person}{T. Jebara}.}
  \bibinfo{year}{2018}\natexlab{b}.
\newblock \showarticletitle{Variational autoencoders for collaborative
  filtering}. In \bibinfo{booktitle}{{\em Proc. of the 2018 World Wide Web
  Conference}} {\em (\bibinfo{series}{WWW '18})}. International World Wide Web
  Conferences Steering Committee, \bibinfo{publisher}{ACM},
  \bibinfo{pages}{689--698}.
\newblock


\bibitem[\protect\citeauthoryear{Lumbreras, Filstroff, and
  F{\'e}votte}{Lumbreras et~al\mbox{.}}{2018}]%
        {lumbreras2018bayesian}
\bibfield{author}{\bibinfo{person}{Alberto Lumbreras}, \bibinfo{person}{Louis
  Filstroff}, {and} \bibinfo{person}{C{\'e}dric F{\'e}votte}.}
  \bibinfo{year}{2018}\natexlab{}.
\newblock \showarticletitle{Bayesian mean-parameterized nonnegative binary
  matrix factorization}.
\newblock \bibinfo{journal}{{\em arXiv preprint arXiv:1812.06866\/}}
  (\bibinfo{year}{2018}).
\newblock


\bibitem[\protect\citeauthoryear{Malkov and Yashunin}{Malkov and
  Yashunin}{2018}]%
        {malkov2018efficient}
\bibfield{author}{\bibinfo{person}{Yury~A Malkov} {and}
  \bibinfo{person}{Dmitry~A Yashunin}.} \bibinfo{year}{2018}\natexlab{}.
\newblock \showarticletitle{Efficient and robust approximate nearest neighbor
  search using hierarchical navigable small world graphs}.
\newblock \bibinfo{journal}{{\em IEEE transactions on pattern analysis and
  machine intelligence\/}} (\bibinfo{year}{2018}).
\newblock


\bibitem[\protect\citeauthoryear{Mikolov, Sutskever, Chen, Corrado, and
  Dean}{Mikolov et~al\mbox{.}}{2013}]%
        {mikolov2013efficient}
\bibfield{author}{\bibinfo{person}{Tomas Mikolov}, \bibinfo{person}{Ilya
  Sutskever}, \bibinfo{person}{Kai Chen}, \bibinfo{person}{Greg~S Corrado},
  {and} \bibinfo{person}{Jeff Dean}.} \bibinfo{year}{2013}\natexlab{}.
\newblock \showarticletitle{Distributed Representations of Words and Phrases
  and their Compositionality}.
\newblock In \bibinfo{booktitle}{{\em Advances in Neural Information Processing
  Systems 26}}, \bibfield{editor}{\bibinfo{person}{C.~J.~C. Burges},
  \bibinfo{person}{L.~Bottou}, \bibinfo{person}{M.~Welling},
  \bibinfo{person}{Z.~Ghahramani}, {and} \bibinfo{person}{K.~Q. Weinberger}}
  (Eds.). \bibinfo{publisher}{Curran Associates, Inc.},
  \bibinfo{pages}{3111--3119}.
\newblock


\bibitem[\protect\citeauthoryear{Neal}{Neal}{2012}]%
        {neal2012bayesian}
\bibfield{author}{\bibinfo{person}{Radford~M Neal}.}
  \bibinfo{year}{2012}\natexlab{}.
\newblock \bibinfo{booktitle}{{\em Bayesian learning for neural networks}}.
  Vol.~\bibinfo{volume}{118}.
\newblock \bibinfo{publisher}{Springer Science \& Business Media}.
\newblock


\bibitem[\protect\citeauthoryear{Nolan and Wand}{Nolan and Wand}{2017}]%
        {nolan2017accurate}
\bibfield{author}{\bibinfo{person}{Tui~H Nolan} {and} \bibinfo{person}{Matt~P
  Wand}.} \bibinfo{year}{2017}\natexlab{}.
\newblock \showarticletitle{Accurate logistic variational message passing:
  algebraic and numerical details}.
\newblock \bibinfo{journal}{{\em Stat\/}} \bibinfo{volume}{6},
  \bibinfo{number}{1} (\bibinfo{year}{2017}), \bibinfo{pages}{102--112}.
\newblock


\bibitem[\protect\citeauthoryear{Paszke, Gross, Chintala, Chanan, Yang, DeVito,
  Lin, Desmaison, Antiga, and Lerer}{Paszke et~al\mbox{.}}{2017}]%
        {paszke2017automatic}
\bibfield{author}{\bibinfo{person}{Adam Paszke}, \bibinfo{person}{Sam Gross},
  \bibinfo{person}{Soumith Chintala}, \bibinfo{person}{Gregory Chanan},
  \bibinfo{person}{Edward Yang}, \bibinfo{person}{Zachary DeVito},
  \bibinfo{person}{Zeming Lin}, \bibinfo{person}{Alban Desmaison},
  \bibinfo{person}{Luca Antiga}, {and} \bibinfo{person}{Adam Lerer}.}
  \bibinfo{year}{2017}\natexlab{}.
\newblock \showarticletitle{Automatic differentiation in PyTorch}. In
  \bibinfo{booktitle}{{\em NIPS-W}}.
\newblock


\bibitem[\protect\citeauthoryear{Ritov, Bickel, Gamst, Kleijn,
  et~al\mbox{.}}{Ritov et~al\mbox{.}}{2014}]%
        {ritov2014bayesian}
\bibfield{author}{\bibinfo{person}{Ya'acov Ritov}, \bibinfo{person}{Peter~J
  Bickel}, \bibinfo{person}{Anthony~C Gamst}, \bibinfo{person}{Bastiaan
  Jan~Korneel Kleijn}, {and} \bibinfo{person}{others}.}
  \bibinfo{year}{2014}\natexlab{}.
\newblock \showarticletitle{The Bayesian Analysis of Complex, High-Dimensional
  Models: Can It Be CODA?}
\newblock \bibinfo{journal}{{\it Statist. Sci.}} \bibinfo{volume}{29},
  \bibinfo{number}{4} (\bibinfo{year}{2014}), \bibinfo{pages}{619--639}.
\newblock


\bibitem[\protect\citeauthoryear{Robbins and Monro}{Robbins and Monro}{1951}]%
        {robbins1951stochastic}
\bibfield{author}{\bibinfo{person}{Herbert Robbins} {and}
  \bibinfo{person}{Sutton Monro}.} \bibinfo{year}{1951}\natexlab{}.
\newblock \showarticletitle{A stochastic approximation method}.
\newblock \bibinfo{journal}{{\em The annals of mathematical statistics\/}}
  \bibinfo{volume}{22}, \bibinfo{number}{3} (\bibinfo{year}{1951}),
  \bibinfo{pages}{400--407}.
\newblock


\bibitem[\protect\citeauthoryear{Rohde, Bonner, Dunlop, Vasile, and
  Karatzoglou}{Rohde et~al\mbox{.}}{2018}]%
        {rohde2018recogym}
\bibfield{author}{\bibinfo{person}{David Rohde}, \bibinfo{person}{Stephen
  Bonner}, \bibinfo{person}{Travis Dunlop}, \bibinfo{person}{Flavian Vasile},
  {and} \bibinfo{person}{Alexandros Karatzoglou}.}
  \bibinfo{year}{2018}\natexlab{}.
\newblock \showarticletitle{RecoGym: A Reinforcement Learning Environment for
  the problem of Product Recommendation in Online Advertising}. In
  \bibinfo{booktitle}{{\em REVEAL workshop, ACM Conference on Recommender
  Systems 2018}}.
\newblock


\bibitem[\protect\citeauthoryear{Rohde and Wand}{Rohde and Wand}{2016}]%
        {rohde2016semiparametric}
\bibfield{author}{\bibinfo{person}{David Rohde} {and} \bibinfo{person}{Matt~P
  Wand}.} \bibinfo{year}{2016}\natexlab{}.
\newblock \showarticletitle{Semiparametric mean field variational Bayes:
  General principles and numerical issues}.
\newblock \bibinfo{journal}{{\em The Journal of Machine Learning Research\/}}
  \bibinfo{volume}{17}, \bibinfo{number}{1} (\bibinfo{year}{2016}),
  \bibinfo{pages}{5975--6021}.
\newblock


\bibitem[\protect\citeauthoryear{Ruiz, Titsias, Dieng, and Blei}{Ruiz
  et~al\mbox{.}}{2018}]%
        {ruiz2018augmentold}
\bibfield{author}{\bibinfo{person}{Francisco~JR Ruiz},
  \bibinfo{person}{Michalis~K Titsias}, \bibinfo{person}{Adji~B Dieng}, {and}
  \bibinfo{person}{David~M Blei}.} \bibinfo{year}{2018}\natexlab{}.
\newblock \showarticletitle{Augment and reduce: Stochastic inference for large
  categorical distributions}.
\newblock \bibinfo{journal}{{\em arXiv preprint arXiv:1802.04220\/}}
  (\bibinfo{year}{2018}).
\newblock


\bibitem[\protect\citeauthoryear{Schnabel, Swaminathan, Singh, Chandak, and
  Joachims}{Schnabel et~al\mbox{.}}{2016}]%
        {schnabel2016recommendations}
\bibfield{author}{\bibinfo{person}{Tobias Schnabel}, \bibinfo{person}{Adith
  Swaminathan}, \bibinfo{person}{Ashudeep Singh}, \bibinfo{person}{Navin
  Chandak}, {and} \bibinfo{person}{Thorsten Joachims}.}
  \bibinfo{year}{2016}\natexlab{}.
\newblock \showarticletitle{Recommendations As Treatments: Debiasing Learning
  and Evaluation}. In \bibinfo{booktitle}{{\em Proceedings of the 33rd
  International Conference on International Conference on Machine Learning -
  Volume 48}} {\em (\bibinfo{series}{ICML'16})}. \bibinfo{pages}{1670--1679}.
\newblock


\bibitem[\protect\citeauthoryear{Storkey}{Storkey}{2009}]%
        {Storkey2009}
\bibfield{author}{\bibinfo{person}{A. Storkey}.}
  \bibinfo{year}{2009}\natexlab{}.
\newblock \showarticletitle{When training and test sets are different:
  characterizing learning transfer}.
\newblock \bibinfo{journal}{{\em Dataset shift in machine learning\/}}
  (\bibinfo{year}{2009}), \bibinfo{pages}{3--28}.
\newblock


\bibitem[\protect\citeauthoryear{Swaminathan and Joachims}{Swaminathan and
  Joachims}{2015a}]%
        {swaminathan2015batch}
\bibfield{author}{\bibinfo{person}{Adith Swaminathan} {and}
  \bibinfo{person}{Thorsten Joachims}.} \bibinfo{year}{2015}\natexlab{a}.
\newblock \showarticletitle{Batch learning from logged bandit feedback through
  counterfactual risk minimization.}
\newblock \bibinfo{journal}{{\em Journal of Machine Learning Research\/}}
  \bibinfo{volume}{16} (\bibinfo{year}{2015}), \bibinfo{pages}{1731--1755}.
\newblock


\bibitem[\protect\citeauthoryear{Swaminathan and Joachims}{Swaminathan and
  Joachims}{2015b}]%
        {Swaminathan2015JMLR}
\bibfield{author}{\bibinfo{person}{A. Swaminathan} {and} \bibinfo{person}{T.
  Joachims}.} \bibinfo{year}{2015}\natexlab{b}.
\newblock \showarticletitle{Batch learning from logged bandit feedback through
  counterfactual risk minimization.}
\newblock \bibinfo{journal}{{\em Journal of Machine Learning Research\/}}
  \bibinfo{volume}{16}, \bibinfo{number}{1} (\bibinfo{year}{2015}),
  \bibinfo{pages}{1731--1755}.
\newblock


\bibitem[\protect\citeauthoryear{Titsias}{Titsias}{2016}]%
        {aueb2016one}
\bibfield{author}{\bibinfo{person}{Michalis Titsias}.}
  \bibinfo{year}{2016}\natexlab{}.
\newblock \showarticletitle{One-vs-each approximation to softmax for scalable
  estimation of probabilities}. In \bibinfo{booktitle}{{\em Advances in Neural
  Information Processing Systems}}. \bibinfo{pages}{4161--4169}.
\newblock


\bibitem[\protect\citeauthoryear{Vapnik}{Vapnik}{2013}]%
        {vapnik2013nature}
\bibfield{author}{\bibinfo{person}{Vladimir Vapnik}.}
  \bibinfo{year}{2013}\natexlab{}.
\newblock \bibinfo{booktitle}{{\em The nature of statistical learning theory}}.
\newblock \bibinfo{publisher}{Springer science \& business media}.
\newblock


\end{thebibliography}

\newpage
\section{Supplementary Material}

\subsection{Approximating expectations under the log softmax}

The variational lower bound of BLO (and BLOB) contains a log softmax term.  An alternative to using the re-parameterization trick is to use The Bouchard bound which removes the need for Monte Carlo methods.  The Bouchard bound introduces a further approximation and additional variational parameters $a,\xi$ but produces an analytical bound:
\vspace{-0.1cm}
\begin{align*}
& \mathcal{L} \ge  \mathcal{L}_{\rm Bouch}  = \left( \sum_t^T \bPsi_{v_t} \bmu_{q_\omega}  + \brho_{v_t} \right) \\
  & -T [ 
  a + \sum_p^P \frac{\bPsi_p \bmu_{q_\omega} +\brho_p - a - \xi_p}{2} + \log(1 + e^{\xi_p}) \\
  & + \lambda_{\rm JJ}(\xi_p) 
    \{(\bPsi_p \bmu_{q_\omega}+\brho_p-a)^2 + \bPsi_p \bSigma_{q_\omega} \bPsi_p^T  - \xi_p^2  \}
  ]\\  
  &  - \frac{K}{2} \log( 2 \pi )-  \frac{1}{2}  \{ \bmu_{q_\omega}^T \bmu_{q_\omega} +
  {\rm trace} (\bSigma_{q_\omega})  \} + \frac{1}{2} \log |2 \pi e \bSigma_{q_\omega} |.
\end{align*}

Because the Bouchard bound causes the softmax to decompose into a sum we can avoid the expensive normalization by subsampling some of the terms in the softmax.
\vspace{-0.1cm}
\begin{align*}
& \hat{\mathcal{L}}_{\rm Bouch}(v_{1},...,v_{T} ,n_1,...n_S,\Xi,\bPsi)   = \left( \sum_t^T \bPsi_{v_t} \bmu_{q_\omega}  + \brho_{v_t} \right)\\
&  -T [ 
    a + \frac{P}{S} \sum_{s'=1}^S  \frac{\bPsi_{n_{s'}} \bmu_{q_\omega} +\brho_{n_{s'}} - a - \xi_{n_{s'}} }{2} + \log(1 + e^{\xi_{n_{s'}}})\\
    & + \lambda_{\rm JJ}(\xi_{n_{s'}}) \times \{(\bPsi_{n_{s'}} \bmu_{q_\omega}+\brho_{n_{s'}}-a)^2 + \bPsi_{n_{s'}} \bSigma_{q_\omega} \bPsi_{n_{s'}}^T  - \xi_{n_{s'}}^2  \}
    ]\\  
    &  - \frac{K}{2} \log( 2 \pi )-  \frac{1}{2}  \{ \bmu_{q_\omega}^T \bmu_{q_\omega} +
    {\rm trace} (\bSigma_{q_\omega})  \} + \frac{1}{2} \log |2 \pi e \bSigma_{q_\omega} |.
  \end{align*}

\noindent
where $v_{1},...,v_{T}$ are the items associated with the session and $n_1,...n_S$ are $S<P$ negative items randomly sampled, and $\lambda_{\rm JJ}(\cdot)$ is the Jaakola and Jordan function \cite{jaakkola1997variational}:

  \[
  \lambda_{\rm JJ}(\xi) = \frac{1}{2\xi} \left( \frac{1}{1+e^{-\xi}} - \frac{1}{2} \right).
  \]

This algorithm is similar to the word2vec algorithm \cite{mikolov2013efficient} but without any non-probabilistic heuristics.

\subsection{Log concavity bound}

The log concave bound \cite{ruiz2018augmentold,lafferty2006correlatedold,bouchard2007efficient} also breaks the log softmax into a sum
\begin{align*}
\log & ~   p(v_1,..,v_T,\bomega_u|\bPsi)  = \left( \sum_t^T \bPsi_{v_t}
            \bomega_u + \brho_{v_t}\right) \\
&  -T \log\{ \sum_p^P \exp( \bPsi_p \bomega_u  +\brho_p)\}  - \frac{K}{2} \log( 2 \pi )-
                                       \frac{1}{2}  \bomega_{u}^T
                                       \bomega_{u} \\
& \ge \left( \sum_t^T \bPsi_{v_t}
\bomega_u + \brho_{v_t}\right) \\
&  -T \phi \{ \sum_p^P \exp( \bPsi_p \bomega_u  +\brho_p)\} + T \log \phi + T - \frac{K}{2} \log( 2 \pi )-
                           \frac{1}{2}  \bomega_{u}^T
                           \bomega_{u} \\
& = L_{\rm log}
\end{align*}

\begin{align*}
&\mathcal{L}_{log} =  E_{q(\omega)}[L_{\rm log}] - {\rm KL}(Q,P) = \left( \sum_t^T \bPsi_{v_t}
\bmu_{q_\omega} + \brho_{v_t}\right) \\
&  -T \phi \{ \sum_p^P \exp( \bPsi_p \bmu_{q_\omega}  +\brho_p + \frac{1}{2} \bPsi_p \bSigma_{q_\omega} \bPsi_p^T )\} + \log \phi + 1  \\
&    - {\rm KL}(Q,P).
\end{align*}

\noindent
A fast noisy version of the bound is:

\begin{align*}
  &\hat{\mathcal{L}}_{log}(v_1,..,v_T,n_1,n_{S_{{\rm neg}}} ) = \left( \sum_t^T \bPsi_{v_t}
  \bmu_{q_\omega} + \brho_{v_t}\right) - {\rm KL}(Q,P) \\
  &  -T \frac{P}{S_{\rm neg}} \phi \{ \sum_{s'}^{S_{\rm neg}} \exp( \bPsi_{n_{s'}} \bmu_{q_\omega}  +\brho_{n_{s'}} + \frac{1}{2} \bPsi_{n_{s'}} \bSigma_{q_\omega} \bPsi_{n_{s'}}^T )\} + T  \log \phi + T
  \end{align*}

\noindent
Finally the one vs each bound \cite{aueb2016one} also breaks the log softmax into a sum without introducing any variational parameter whatsoever.

We can also use a variational auoto-encoders for $a, \xi$ in the case of the Bouchard bound and
$\phi$ in the case of the log concave bound to prevent variational parameters growing with the size of the dataset.  This is similar to the augment and reduce approach \cite{ruiz2018augmentold} but has no requirement to be in complete data exponential family form.

The computational impact of turning the log softmax into a sum computationally is driven by $P$ and GPU size.  If $P$ is small compared to the GPU it may be preferable to avoid using any additional approximations and compute the full softmax using the re-parameterization trick.

\subsection{The EM Algorithm - an alternative to the VAE}
\subsubsection{Standard EM algorithm}
  If the parameters $\bPsi,\brho$ are already known then the posterior over the user embedding $\bomega$ may be calculated by optimizing the lower bound using the following variational EM algorithm. The EM algorithm exploits the fact that the Bouchard bound is quadratic and conjugate to the Gaussian distribution.
  The algorithm here is the \emph{dual} of the one presented in \cite{bouchard2007efficient} as we assume the embedding $\bPsi$ is fixed and $\bomega$ is updated where the algorithm they present does the opposite.  The EM algorithm consists of cycling the following update equations:

\[
\bSigma_{q_\omega}^{-1} = I_k + 2 T \sum_p \lambda_{\rm JJ}(\xi_p)\bPsi_p^T \bPsi_p,
\]

\[
\bmu_{q_\omega} = \bSigma_{q_\omega}  \left(  ( \sum_t^T \bPsi_{v_t}^T) - T \left[ \sum_p^P
  \{ \frac{1}{2} + 2(\brho_p-a) \lambda_{\rm JJ}(\xi_p) \} \bPsi_p^T  \right] \right),
\]

\[
a = \frac{-1+\frac{P}{2} + \sum_p  2 \lambda_{\rm JJ}(\xi_p) (\bPsi_p \bmu_{q_\omega} + \brho_p) }{ 2 \sum_p \lambda_{\rm JJ}(\xi_p) },
\]

\[
\xi_{p} = h(\bPsi_p,\brho_p,a,\bSigma_{q_\omega},\brho_q) = \sqrt{\bPsi_p \bSigma_{q_\omega} \bPsi_p^T + (\bPsi_p \bmu_{q_\omega} + \brho_p-a)^2 }.
\]

\subsubsection{Fast online EM algorithm}
We further note that the EM algorithm is (with the exception of the $a$ variational parameter) a fixed point update (of the natural parameters) that decomposes into a sum.  The terms in the sum come from the softmax in the denominator.  After substituting a co-ordinate descent update of $a$ with a gradient descent step update, then the entire fixed point update becomes a sum:

\[
(\bSigma_{q_\omega}^{-1})^{\rm new} = I_k + 2 \sum_p \lambda_{\rm JJ}(
  h(\bPsi_p,\brho_p,a,\bSigma_{q_\omega},\brho_q)
 )\bPsi_p^T \bPsi_p,
\]

\begin{align*}
&  (\bSigma_{q_\omega}^{-1} \bmu_{q_\omega})^{\rm new} =   ( \sum_t^T \bPsi_{v_t}^T) \\
&  - T \left[ \sum_p^P
  \{ \frac{1}{2} + 2(\brho_p-a) \lambda_{\rm JJ}\{
    h(\bPsi_p,\brho_p,a,\bSigma_{q_\omega},\brho_q)   \} \} \bPsi_p^T  \right] 
\end{align*}

\begin{align*}
a^{\rm new}  = & a + \frac{-1+\frac{P}{2}}{2} \\
& + \sum_p   \lambda_{\rm JJ}\{
  h(\bPsi_p,\brho_p,a,\bSigma_{q_\omega},\brho_q) 
\} \\
& \times  (\bPsi_p \bmu_{q_\omega} + \brho_p) - a \lambda_{\rm JJ}\{  h(\bPsi_p,\brho_p,a,\bSigma_{q_\omega},\brho_q) \}
\end{align*}

\noindent
That is the EM algorithm can be written: 

\[
\left( (\bSigma_{q_\omega}^{-1})^{\rm new}, (\bSigma_{q_\omega}^{-1} \bmu_{q_\omega})^{\rm new}, a^{\rm new} \right) = \sum_p^P g(\bPsi_p,\brho_p,\bSigma_{q_\omega}^{-1}, \bSigma_{q_\omega}^{-1} \bmu_{q_\omega}, a).
\]

\noindent
As noted in \cite{cappe2009line} when an EM algorithm can be written as a fixed point update over a sum, then the Robbins Monro algorithm can be applied.  Allowing updates of the form ($p$ is chosen randomly):

\begin{align*}
 (\bSigma_{q_\omega}^{-1})^{\rm (s)},& (\bSigma_{q_\omega}^{-1} \bmu_{q_\omega})^{\rm (s)}, a^{\rm (s)}    \\
&= (1-\Delta_s)\left( (\bSigma_{q_\omega}^{-1})^{\rm (s-1)}, (\bSigma_{q_\omega}^{-1} \bmu_{q_\omega})^{\rm (s-1)}, a^{\rm (s-1)} \right) \\
& +  \Delta_s g(\bPsi_p,\brho_p,(\bSigma_{q_\omega}^{-1})^{\rm (s-1)}, (\bSigma_{q_\omega}^{-1} \bmu_{q_\omega})^{\rm (s-1)}, a^{\rm (s-1)}).
\end{align*}

\noindent
where $\Delta$ is a slowly decaying Robbins Monro sequence (\cite{robbins1951stochastic}) with $\Delta_1=1$ (meaning no initial value of $(\bSigma_{q_\omega}^{-1})^{\rm (0)}, (\bSigma_{q_\omega}^{-1} \bmu_{q_\omega})^{\rm (0)}, a^{\rm (0)}$) is needed.  For large $P$ this algorithm is many times faster than the generic EM algorithm.  Note that (unusually) the Robbins Monro algorithm is applied to the softmax of a large categorical variable and not to individual records under a conditionally independent assumption.

There are other variational bounds that may be considered for this problem most notably the tilted bound \cite{knowles2011non}.  For the tilted bound the known fixed point algorithms are not guaranteed to be stable and are not always stable in practice \cite{nolan2017accurate,rohde2016semiparametric} so extra methods such as line searches would need to be considered.the tilted bound also does not decompose into a sum.  We do not further consider alternative bounds.

The computational cost of this algorithm depends on the number of products $P$ linearly and the embedding size $K$ cubicly, if $P$ and $K$ are modest it can take less than a second making it potentially deployable at prediction time.  In practice we found the cost of large $P$ might be prohibitive due to the sums over all $P$ embeddings, in these cases a variational auto-encode described in the next section, is to be preferred.

\subsection{Next Item Prediction}
The predictive distribution required to do next item prediction is also not trivial in this case, i.e. approximating:

\begin{align*}
&  p(v_{u,T+1}|v_{u,1},..,v_{u,T}) \\
&  = \int  p(v_{u,T+1}|\bomega,\bPsi,\brho) p(\bomega|v_{u,1},.v_{u,T})  d \bomega_u
\end{align*}

\noindent
is not trivial even if $p(\bomega|v_{u,1},..v_{u,T_u})$ is approximated with a Gaussian distribution $\bomega_u|v_1,..v_T \sim \mathcal{N}(\bmu_{q_\omega},\bSigma_{q_\omega})$.  We are interested in computing:

\[
 p(v_{n+1}|v_1,...v_n) \approx \E_{q(\bomega)} \left[ \frac{\exp(\bPsi_v \bomega + \brho) }{\sum_{v'} \exp(\bPsi_{v'} \bomega + \brho)  }\right].
\]

\noindent
We considered using a Monte Carlo based approximation, first by drawing $S$ samples:
\[
\bomega^{(s)}  \sim \mathcal{N}(\bmu_{q_\omega}, \bSigma_{q_\omega}),
\]

\[
  p(v_{n+1}|v_1,...v_n) \approx \frac{1}{S} \sum_s^S \frac{\exp(\bPsi_v \bomega^{(s)} + \brho) }{\sum_{v'} \exp(\bPsi_{v'} \bomega^{(s)} + \brho)  },
\]

\noindent
as well as using a number of fast approximations such as:

\[
  p(v_{n+1}|v_1,...v_n) \approx  \frac{\exp(\bPsi_v \bmu_{q_\omega} + \brho) }{\sum_{v'} \exp(\bPsi_{v'} \bmu_{q_\omega} + \brho)  }.
\]

\noindent
while we investigated more complex approximations (such as normalizing the exponential of the lower bound) we did not find they helped in practice.

\end{document}